%% file: iclr2026_conference.tex
\documentclass{article}
\usepackage{iclr2026_conference,times}

\input{math_commands.tex}

\usepackage{hyperref}
\usepackage{xurl}
\usepackage{graphicx}
\usepackage{colortbl}
\usepackage{booktabs}
\usepackage{multirow}
\usepackage{enumitem}
\usepackage{wrapfig}
\usepackage{pifont}
\usepackage[capitalise]{cleveref}
\definecolor{color5}{HTML}{006795}
\hypersetup{
  colorlinks   = true, %Colours links instead of ugly boxes
  urlcolor     = blue, %Colour for external hyperlinks
  linkcolor    = color5, %Colour of internal links
  citecolor   = color5 %Colour of citations, could be ``red''
}
\usepackage[framemethod=TikZ]{mdframed}
\definecolor{UserExampleBg}{HTML}{ffffff}
\definecolor{UserExampleTitle}{HTML}{618197}
\newmdenv[
    roundcorner=5pt,
    backgroundcolor=UserExampleBg,
    linecolor=UserExampleTitle,
    outerlinewidth=0.3pt,
    frametitlebackgroundcolor=UserExampleTitle,
    frametitlefont={\bfseries\color{white}},
]{user_example}

\AtBeginEnvironment{user_example}

\title{Nudging the Boundaries of LLM Reasoning\vspace{10pt}}
\author{
  \hspace{12pt}
  \shortstack{
    Justin Chih-Yao Chen$^{1,2}$ \quad Becky Xiangyu Peng$^{1}$ \quad Prafulla Kumar Choubey$^{1}$ \\[4pt]
    Kung-Hsiang Huang$^{1}$ \quad Jiaxin Zhang$^{1}$ \quad Mohit Bansal$^{2}$ \quad Chien-Sheng Wu$^{1}$
    \\[4pt]
    \normalfont
    $^{1}$Salesforce AI Research \quad
    $^{2}$UNC Chapel Hill \quad
  }
}

\usepackage{xcolor}
\definecolor{brightcerulean}{rgb}{0.11, 0.67, 0.84}

\newcommand{\method}[0]{NuRL}

\newcommand{\cmark}{\ding{51}}%
\newcommand{\xmark}{\ding{55}}%

\iclrfinalcopy % Uncomment for camera-ready version, but NOT for submission.
\begin{document}

\maketitle

\begin{abstract}
Current online reinforcement learning (RL) algorithms like GRPO share a key limitation in LLM reasoning: they cannot learn from problems that are ``unsolvable'' to the model. In other words, they can only improve performance on problems where the model is capable of exploring the correct answer. If a problem is too difficult--such that even hundreds of attempts never produce a correct solution--the model cannot learn from it. Consequently, the model's ``upper limit'' remains unchanged after RL training, even though the likelihood of solving easier, solvable problems may increase. These hard, unsolvable samples--though potentially rich in learning signal--cannot contribute to training, as no rollouts yield rewards and thus no gradients are produced.
To unlock learning from these hard samples, we propose \method{}
\footnote{Pronounced like “neural” (nur·uhl)}, a ``nudging'' method that aims to push the upper bound of LLM reasoning using self-generated hints, i.e., abstract cues that help reduce the problem difficulty for the model. Given a question and its gold answer, the model generates a Chain-of-Thought (CoT) and then produces a hint containing the core knowledge needed to solve the problem.
During online RL training, we generate $\mathcal{G}$ rollouts from the base policy and use the pass rate to decide whether the hint should be injected.
For hard samples with a 0\% pass rate, we inject the offline-generated hint and regenerate a new batch of trajectories. This yields two benefits: (1) the hint boosts pass rates (from 0\% to non-zero), thereby introducing training signals for previously unsolvable samples, and (2) the hints are self-generated (conditioned on the gold answer), avoiding distributional shift and do not rely on external models.
Compared to standard GRPO, \method{} achieves consistent improvements across six diverse benchmarks and three models, while remaining complementary to test-time scaling. Notably, \method{} can raise the model's upper limit, whereas GRPO leaves pass@1024 unchanged from the base model. Furthermore, we present a systematic study of what makes an effective hint and when hints are most useful. Interestingly, the best hints are abstract and high-level--as revealing gold answers actually hurt performance--and are most beneficial when applied necessarily and after GRPO has converged.\footnote{Code: \url{https://github.com/SalesforceAIResearch/NuRL}}
\end{abstract}

\section{Introduction}
Recent advances in reinforcement learning (RL) algorithms have played a central role in improving the reasoning abilities of large language models (LLMs). Despite many promising advances, current online RL algorithms share a key limitation: they cannot learn from problems that are unsolvable under the base policy. In other words, if the model cannot reach the correct answer even after extensive exploration, then no meaningful learning signal can be obtained from the problem.
On a similar vein, a growing body of work finds that post-training mainly encourages models to generate already high-reward trajectories~\citep{he2025rewarding,yue2025does, dang2025assessing, zhao2025echo}. 
As a result, the model's upper limit--often measured by pass@$k$ for large $k$—remains unchanged after RL training.
Intuitively, learning from harder samples offers a clear path to improving a model's performance and expanding its ceiling capacity. In other words, learning from the hard samples has two key benefits: (1) extracting more training signal from the same dataset (improving pass@1), and (2) enabling the model to solve previously unsolvable problems, thereby extending its capability boundary (improving pass@$k$). However, precisely because these problems are difficult, models often cannot learn them without appropriate guidance or intervention from a teacher model.

\begin{figure*}
    \centering
    \includegraphics[width=\linewidth]{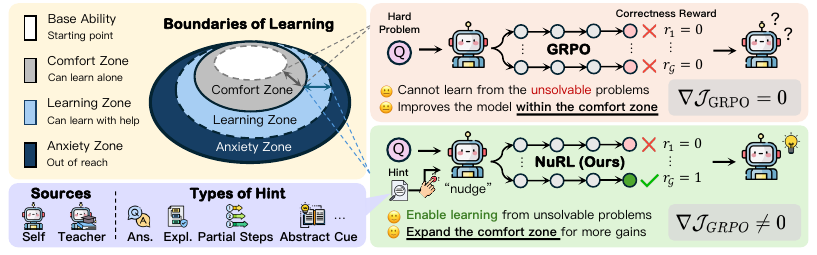}
    \vspace{-18pt}
    \caption{GRPO yields substantial gains, but the improvements largely stem from extending the model's ability within its comfort zone, i.e., if the model fails to solve a hard problem after numerous attempts, it is unable to learn from that problem. In \method{}, we address this by exploring various forms of hints (abstract cues, partial steps, explanations, or even the gold answer), which can be self-generated or teacher-generated. Both self- and teacher-generated abstract cues can expand the model's comfort zone, effectively transforming previously unsolvable problems into solvable ones. 
    }
    \label{fig:fig1}
\end{figure*}

\begin{figure*}[t!]
    \centering
    \includegraphics[width=1.0\textwidth]{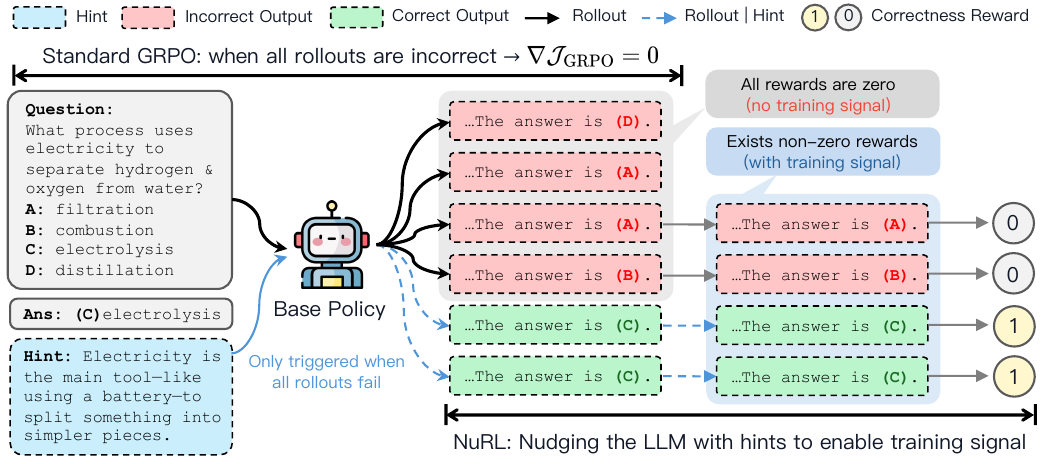}
    \caption{\method{} provides targeted guidance to the LLM policy during online GRPO training. Prior to training, we construct an offline collection of hints, defined as abstract problem-specific cues that reduce task difficulty. During the online training, whenever all $\mathcal{G}$ rollouts for a problem are incorrect, \method{} augments $\mathcal{G}-1$ of the rollouts with the corresponding hint and regenerates the batch. This intervention facilitates the acquisition of non-zero rewards on instances that would otherwise yield uniformly zero rewards, thereby supplying informative training signals.
    }
    \label{fig:method}
\end{figure*}

This dynamic parallels Vygotsky’s concept of the Zone of Proximal Development (ZPD)~\citep{vygotsky1978development}, which distinguishes between tasks a learner can solve independently (“comfort zone” in \cref{fig:fig1}) and those achievable only with appropriate guidance (“learning zone” in \cref{fig:fig1}). The inability to learn from hard samples -- or the lack of improvement in pass@$k$ -- mirrors being trapped in the comfort zone. Motivated by this analogy, we ask: \textit{can models generate their own ``hints""—lightweight forms of guidance—so that even the hard problems become learnable?}
Then, we propose \method{} (\textbf{Nu}dging LLM with \textbf{R}einforcement \textbf{L}earning), which adaptively injects self-generated hints into training. 
Our hypothesis is that hard problems become more learnable when paired with carefully abstracted hints~\citep{huang2025boosting,park2025deepvideo}. These hints act as light-weight ``nudges,'' transforming previously unlearnable samples into productive training signals.
To achieve this, our approach begins with offline hint collection. Given a question and its gold answer, we prompt the model to generate Chain-of-Thought \citep[CoT;][]{wei2022chain} reasoning that connects the two. Using the question and CoT as input, the model then produces a high-level cue that captures the core knowledge required to solve the problem. 
We also explore various types of hints as shown in the bottom-left of \cref{fig:fig1}, where the hints can be self-generated or provided by a stronger model (e.g., GPT-o4-mini \citep{gpto4mini}), and the forms of hints can be abstract cues, partial steps, explanations, or even the gold answer.
We find that self-generated hints are effective, while teacher-generated hints give further improvements. 
Importantly, effective hints are abstract and conceptual: they neither reveal the final answer nor provide detailed solution steps, but only mention what core knowledge is needed to solve this problem (see the bottom left in \cref{fig:method}). 

We adopt GRPO~\citep{shao2024deepseekmath} as the training framework. During training, the policy model generates $\mathcal{G}$ rollouts per problem, and we use pass rate to decide when to inject hints. For hard problems with a 0\% pass rate, we inject the pre-generated hints to the end of the question, and prompt the model to regenerate another $\mathcal{G}$ rollouts conditioned on the hints. Given the hint, the model is more likely to produce successful solutions (non-zero pass rate), turning previously unlearnable examples into learnable ones.
This strategy offers two main advantages:
(1) Hints boost pass rate, enabling hard problems to produce meaningful training signals. 
(2) Hints remain within the model's distribution, since they are self-generated (conditioned on the answer), avoiding distributional shift and does not require stronger external models.
Together, these benefits allow \method{} to unlock value from harder samples, broadening the set of problems that contribute to RL training. 

We evaluate \method{} on six diverse benchmarks across multiple domains, including MATH 500, MATH Hard, AIME, GPQA, MMLU-Pro and Date Understanding. Results show that \method{} boosts the average performance with three different models ($+1.62\%$ using Llama, $+1.75\%$ using OctoThinker, and $+0.79\%$ using Qwen as compared to GRPO), and when an external teacher model is available for hint generation, the improvement can be further enlarged to up to $3.44\%$.
Moreover, our analysis shows that 
(a) \method{} is complementary to test-time scaling methods such as Self-Consistency \citep{wang2022self} and shows larger improvement: our method improves $9.4\%$ with 16-way Self-Consistency, as compared to GRPO, which improves $7.8\%$.
(b) \method{} is able to transform previously unsolvable problems into solvable ones, and that transfer to the improvements on the upper limit of a model's capacity.
(c) Hints are useful when they are abstract and high-level. \textit{The more exposure to the answer, the more severe the degradation}. This aligns with human learning, where effective hints should be abstract and high-level—providing guidance without revealing the solution. Knowing the answer upfront risks biasing toward it and undermines generalization.

\section{Related Work}
\textbf{Reinforcement Learning with Verifiable Reward (RLVR).}
RLVR computes the reward using rule-based verification, which is effective in improving LLM reasoning. The reward function can be as simple as checking whether the model's answer matches the gold answer \citep{lambert2024tulu, guo2025deepseek, team2025kimi, zeng2025simplerl}. 
The success of RLVR is also supported by advancements, including PPO \citep{schulman2017proximal}, DPO \citep{rafailov2023direct}, GRPO \citep{guo2025deepseek} and many techniques like DAPO \citep{yu2025dapo}, Dr. GRPO \citep{liu2025understanding}. Recent work explores different strategies to learn a better self-reflection ability, such as ``try again'' \citep{liu2025simple} or self-critique \citep{zhang2025critique}.

\noindent\textbf{The Role of RL: Distribution Sharpening vs. Discovery.}
There has been an active discussion on whether RL primarily performs \emph{distribution sharpening}, i.e., amplifying behaviors already present in the model, or enables genuine discovery of new reasoning abilities. The distribution-sharpening view holds that RL mainly surfaces high-reward paths and increases their likelihood of generation \citep{zhang2025rlsurvey, zhao2025echo, shenfeld2025rl}. This is often supported by the findings that RL improves pass@1 but not pass@$k$ \citep{yue2025does, he2025rewarding}, and that even weak reward signals can yield substantial improvements \citep{shao2025spurious, zuo2025ttrl, prabhudesai2025maximizing, wang2025reinforcement}.
On the other hand, some argue that RL fails to improve pass@$k$ not because of inherent limitations, but due to insufficient training or evaluating on tasks where models already perform strongly \citep{liu2025prorl}. \cite{yuan2025llms} also show that RL can learn new skills by combining existing ones. 
\cite{zhang2025rlsurvey} suggest that RL can both sharpen and discover, with the balance determined by the trade-off between exploration and exploitation.

\noindent\textbf{Mixture of Offline and Online RL.}
When the base policy fails to generate solutions that yield non-zero rewards, replay buffer or off-policy optimization leverage previous positive trajectories \citep{lu2025arpo} or expert demonstrations \citep{levine2020offline}. Recently, hybrid approaches that mix online and offline training have been proposed to improve generalization \citep{yan2025learning, phan2025migrate}. 
TBA \citep{bartoldson2025trajectory} decouples data generation from policy updates by leveraging multiple searcher nodes to produce diverse trajectories, enabling high resource utilization and a more scalable search process.
A line of work uses SFT with RL to expand the model's knowledge scope \citep{ma2025learning, fu2025srft, zhou2025expo}, while others employ hints (e.g., partial solutions) to adjust problem difficulty during training \citep{zhang2025stephint, huang2025boosting, li2025questa, qu2025rlad, wang2025dissmath}.
Our work differs from these approaches in two aspects. First, \method{} does not rely on SFT to broaden the model's knowledge; instead, we focus on the RL stage and analyze how different hints expand a model's reasoning boundary. A notable work is STaR \citep{zelikman2022star}, where the reasoning is bootstrapped from reasoning given the answer. Our hints are further abstracted out from the reasoning to ensure they do not disclose the answer. Second, we show that nudging the model with self-generated hints is effective: it enables self-improvement (both pass@1 and pass@$k$) without depending on stronger external models.

\section{Methodology}
\label{sec:method}
\subsection{Preliminary}
Our method is based on GRPO~\citep{shao2024deepseekmath}. 
Specifically, GRPO updates the policy by maximizing $\mathcal{J}_\text{GRPO}(\theta)$ using the following objective:

\begin{equation}
\mathcal{J}_{\text{GRPO}}(\theta) = \frac{1}{\mathcal{G}} \sum_{i=1}^{\mathcal{G}} \frac{1}{|o_i|} \sum_{t=1}^{|o_i|}
\min \left[
\frac{\pi_\theta(o_{i,t} | o_{i,<t})}{\pi_{\text{old}}(o_{i,t} | o_{i,<t})} \hat{A}_{i,t},\,
\text{clip} \left(
\frac{\pi_\theta(o_{i,t} | o_{i,<t})}{\pi_{\text{old}}(o_{i,t} | o_{i,<t})}, 1 - \varepsilon, 1 + \varepsilon
\right) \hat{A}_{i,t}
\right], \notag
\end{equation}

where $\pi_{\theta}$ is the policy, $\pi_{\text{old}}$ is the old policy, $\varepsilon$ is the clipping range, and $\hat{A}_t$ is an estimator of the advantage at time step $t$. Given a reward function $f$ and a question-answer pair $(q,a)$ from training data $\mathcal{D}$, the advantage is estimated by letting $\pi_{\theta_\text{old}}$ samples a group of $\mathcal{G}$ responses $\{ o_i\}_{i=1}^\mathcal{G}$. Then, the advantage of the $i$-th response is calculated by normalizing the rewards within the group $\{ r_i \}_{i=1}^\mathcal{G}$:
\begin{equation}
\hat{A}_{i,t} = \frac{r_i - \mu_r}{\sigma_r}, 
\quad \mu_r = \frac{1}{\mathcal{G}} \sum_{i=1}^{\mathcal{G}} r_i, 
\quad \sigma_r = \sqrt{\frac{1}{\mathcal{G}} \sum_{i=1}^{\mathcal{G}} (r_i - \mu_r)^2}.
\end{equation}

We use the rule-based outcome correctness as the reward \citep{guo2025deepseek}, i.e., $f(\hat{y}, y) = 1, \text{if } \texttt{is\_equivalent}(\hat{y}, y) \text{ else } 0$, where $y$ is the ground truth and $\hat{y}$ is the predicted answer.

\subsection{\method{}: \textbf{Nu}dging LLMs with \textbf{R}einforcement \textbf{L}earning}
\textbf{Offline Hint Collection.}
Given a training dataset consisting of question–answer pairs $\mathcal{D}={(q_i,a_i)}_{i=1}^N$, our method begins with offline hint collection. As illustrated in \cref{fig:fig1}, hints can be categorized by their \emph{source} (self-generated by the model vs.\ provided by a teacher model) and their \emph{type}:
\textbf{(1) Abstract cues}: generated by abstracting from an explanation of why an answer is correct, designed to avoid revealing concrete details or the final answer.
\textbf{(2) Partial steps}: obtained by generating a step-by-step solution using the gold answer, masking 75\% of the steps (keep only the first 25\%), with the model asked to complete the rest.
\textbf{(3) Explanations}: formed by replacing incorrect rollouts with gold-conditioned explanations that justify why the answer is correct.
\textbf{(4) Ground-truth answer}: appending the correct answer alongside the question with the prompt, ``I was told the answer is \{gold\_answer\}. Do not blindly accept it. Take it as a reference and provide your own step-by-step reasoning.''
Note that from (1) to (4) is a reverse order of how much information about the answer is being revealed. 
Later in \cref{sec:hint_type_analysis}, we will show that the more disclosure about the answer, the worse the performance is. 
Therefore, in this work, we mainly focus on \textbf{self-generated abstract cues} as the hint (see \cref{sec:hint_type_examples} for examples with different types of hints).

Concretely, to collect such high-level abstract cues that can simplify the problem without revealing specific solution details, we first prompt the base policy LLM with both the question and the ground-truth answer, and instruct it to generate a Chain-of-Thought \citep[CoT;][]{wei2022chain} that explains why the provided answer is correct.
This can be expressed as $y=\pi_{{\text{old}}}(q,a;p_y)$ where $\pi_{{\text{old}}}$ is the initial policy, $y$ is the generated CoT, and $p_y$ is the prompt (see \cref{sec:prompt}).
We find that directly using such explanation-style CoTs as hints is not effective (also will be shown in \cref{sec:hint_type_analysis}).
To address this, we introduce an abstraction step. Specifically, we prompt the LLM again with the question, the answer, and the self-generated CoT $(q,a,y)$, asking it to produce a high-level hint that reduces task difficulty without being overly specific: $h = \pi_\theta(q,a,y;p_h)$ where $p_h$ is the hint generation prompt as provided in \cref{sec:prompt}. 
After this process, the training data is augmented with hints, yielding an enriched dataset $\mathcal{D}=\{(q_i,a_i,h_i)\}_{i=1}^N$ where every sample has a corresponding hint.

\textbf{Online Rollout Augmentation.}
After augmenting the training data with hints, we proceed with online GRPO-style training. For each question $q$, the base policy first generates $\mathcal{G}$ rollouts without hints. When all responses ${\{o_i\}}_{i=1}^\mathcal{G}$ are incorrect (i.e., $r_i=0 \Rightarrow \hat{A}_{i,t}=0$ for all $i$), the advantages vanish, yielding $\nabla \mathcal{J}_\text{GRPO} = 0$ and thus no policy update. This is illustrated in the gray box of \cref{fig:method}, where the correct answer is C but none of the rollouts reach it. Similarly, if all rollouts are correct, the task is trivially easy and again results in $\nabla \mathcal{J}_\text{GRPO} = 0$. In practice, such uninformative problems (both too hard and too easy) are often discarded to improve training efficiency \citep{yu2025dapo}.
In contrast, \method{} retains the hard cases, as they present opportunities to unlock further gains. Specifically, when all $\mathcal{G}$ rollouts fail, we activate \method{} by appending the offline-generated hint $h$ to the problem, producing the new input $q \oplus h$, where $\oplus$ denotes concatenation. A new batch of rollouts is then generated from $q \oplus h$ (blue box in \cref{fig:method}). To reduce the chance to collapse into uniformly correct responses—which would again eliminate the learning signal—we let only $\mathcal{G}-1$ rollouts see the hint, $\{{o_i}\}_{i=1}^{\mathcal{G}-1} \sim \pi_{\theta_\text{old}}(q,h)$, while one rollout remains hint-free, ${o_\mathcal{G}} \sim \pi_{\theta_\text{old}}(q)$.

\textbf{Inference.}
The hints are only used in training. During test time, we prompt the model only with the question. The hypothesis is that exposure to hints during training guides the model toward correct solutions, enabling it to internalize the reasoning patterns required to solve the problems. That is, the use of hints broadens the set of solvable problems and translates to improved performance.

\section{Experimental Setup}
\textbf{Models.}
We evaluate \method{} on three models: Llama3.2-3B-Instruct \citep{llama3}, OctoThinker-3B-Hybrid-Zero \citep{octo}, and Qwen3-4B-Instruct-2507 \citep{qwen3}.\footnote{Hereafter, we refer to them as Llama, OctoThinker, and Qwen, respectively.}
Llama is a general-purpose instruction-tuned model, while OctoThinker is a recently proposed model that applies \textit{mid-training} on Llama and has been shown to exhibit stronger compatibility with reinforcement learning \citep{octo}. Qwen is also a general-purpose instruction-tuned model that encompasses strong compatibility with post-training techniques.
We evaluate these models because they exhibit distinct properties, helping us avoid conclusions that are overly specific to a single model—a lesson underscored in recent studies \citep{shao2025spurious,llmrl2025incorrect}.

\noindent\textbf{Datasets.}
For training, we use Open-R1's Mixture-of-Thought dataset \citep{openr1} due to its diversity, which includes science QA data from Llama-Nemotron \citep{bercovich2025llamanemotron}, ensuring our training set is not purely math-focused. We randomly sample 7.5k math and 2.5k science data points. Since the dataset provides only CoT outputs from Deepseek-R1 \citep{guo2025deepseek} rather than explicit gold answers, we extract the answers from \textbackslash boxed\{\} and generate 8 CoTs per problem using GPT-o4-mini \citep{gpto4mini}. We keep only samples where Deepseek-R1 and GPT-o4-mini's majority agree on the answer. This yields 8.3k samples for training.
\method{} is tested across a diverse set of reasoning-intensive tasks spanning STEM and other domain-specific areas, including:
\textbf{(1) MATH 500}, a subset 500 problems from the MATH benchmark \citep{hendrycksmath2021} curated by \citet{lightman2023let};
\textbf{(2) MATH Hard}, the hardest problem set from MATH, totaling 1.3k problems \citep{hendrycksmath2021};
\textbf{(3) AIME 2024}, 30 mathematics problems from the 2024 AIME competition \citep{AIME};
\textbf{(4) GPQA Diamond}, 198 PhD-level questions covering biology, physics, and chemistry \citep{rein2024gpqa};
\textbf{(5) MMLU-Pro}, a more challenging variant of MMLU \citep{hendrycks2020measuring}, spanning 14 college-level subjects with 12k samples \citep{wang2024mmlu};
\textbf{(6) Date Understanding}, 250 problems designed to test LLM's understanding to date information, requiring commonsense and logical reasoning \citep{srivastava2023beyond,suzgun2022challenging}.

\noindent\textbf{Baselines.}
We compare \method{} with the following baselines:
(1) \textbf{Zero-shot:} We prompt the model to think step-by-step and provide the answer within \textbackslash boxed{\{}\} \citep{kojima2022large}.
(2) \textbf{Few-shot:} Besides prompting the model to think step-by-step, we include 8 in-context learning samples.
(3) \textbf{Rejection sampling Fine-Tuning (RFT):} We prompt the model with training data 8 times using the zero-shot prompt same as above, and keep only the correct reasoning chains to perform supervise fine-tuning \citep{yuan2023scaling}.
(4) \textbf{Reasoning with Reinforced Fine-Tuning (ReFT):} We adopt the SFTed model from RFT, and continue for GRPO training \citep{trung-etal-2024-reft}.
(5) \textbf{GRPO}: Using outcome correctness as a rule-based reward function \citep{guo2025deepseek}.

\begin{table*}[t]
\vspace{-10pt}
\caption{Comparison of methods across three models and six benchmarks. \method{} consistently outperforms all baselines with self-generated hints, and shows further improvements when an external model is available for hint generation (shown in gray for reference).}
\vspace{6pt}
\centering
\small
\renewcommand{\arraystretch}{1.2}
\begin{tabular}{lccccccc}
\toprule
 & MATH 500 & MATH Hard & AIME & GPQA & MMLU-Pro & Date & Avg. \\
\midrule
\multicolumn{8}{c}{\textbf{Llama3.2-3B-Instruct}} \\
\midrule
Zero-shot                    & 35.71 & 15.28 & 3.33  & 13.23 & 11.52 & 2.33    & 13.57 \\
Few-shot                     & 36.68 & 16.73 & 3.63 & 15.52 & 12.02 & 8.87     & 15.58 \\
RFT                          & 40.12 & 17.72 & 3.63  & 14.76 & 12.98 & 13.36 & 17.10 \\
ReFT                         & 55.80 & 28.86 & 8.00  & 24.42 & 30.19 & 55.34 & 33.77 \\
GRPO                         & 56.92 & 30.11 & 8.33  & 27.98 & 34.78 & 57.10 & 35.87 \\
\rowcolor{blue!7} w/ Hint (Self)               & \textbf{58.04} & \textbf{31.62} & \textbf{9.17}  & \textbf{28.28} & \textbf{36.18} & \textbf{61.65} & \textbf{37.49} \\
\rowcolor{gray!12} w/ Hint (GPT-o4-mini)    & 59.30 & 32.42 & 10.83 & 28.40 & 41.38 & 63.53 & 39.31 \\
\midrule
\multicolumn{8}{c}{\textbf{OctoThinker-3B-Hybrid-Zero}} \\
\midrule
Zero-shot                    & 59.98 & 36.87 & 4.83  & 16.57 & 21.18 & 19.25 & 26.45 \\
Few-shot                     & 61.23 & 37.01 & 5.33 & 18.62 & 23.32 & 24.53 & 28.34 \\
RFT                          & 62.78 & 37.73 & 4.83  & 19.45 & 27.34 & 36.82 & 31.49 \\
ReFT                         & 66.38 & 39.69 & 6.66  & 24.53 & 44.66 & 70.32 & 42.04 \\
GRPO                         & 68.81 & 41.29 & 8.33 & 23.26 & 44.25 & 69.85 & 42.63 \\
\rowcolor{blue!7} w/ Hint (Self)               & \textbf{70.13} & \textbf{42.07} & \textbf{9.66} & \textbf{27.15} & \textbf{45.54} & \textbf{71.75} & \textbf{44.38} \\
\rowcolor{gray!12} w/ Hint (GPT-o4-mini)    & 71.62 & 43.51 & 12.63 & 27.43 & 46.53 & 72.28 & 45.67 \\
\midrule
\multicolumn{8}{c}{\textbf{Qwen3-4B-Instruct-2507}} \\
\midrule
Zero-shot                    & 94.88 & 90.54 & 58.75 & 35.57 & 58.88 & 83.35 & 70.33 \\
Few-shot                     & 94.97 & 90.54 & 55.52 & 34.82 & 59.01 & 85.01 & 69.98 \\
RFT                          & 94.41 & 90.41 & 56.63  & 37.72 & 59.54 & 85.42 & 70.69 \\
ReFT                         & 96.46 & 90.83 & 62.79  & 60.31 & 72.21 & 92.20 & 79.13 \\
GRPO                         & \textbf{96.52} & 90.54 & 60.83 & 62.50 & 72.65 & \textbf{92.80} & 79.31 \\
\rowcolor{blue!7} w/ Hint (Self)               & 96.46 & \textbf{92.57} & \textbf{63.54} & \textbf{62.88} & \textbf{72.83} & 92.30 & \textbf{80.10} \\
\rowcolor{gray!12} w/ Hint (GPT-o4-mini)    & 96.58 & 92.96 & 62.71 & 64.99 & 72.95 & 93.91 & 80.68 \\
\bottomrule
\end{tabular}
\vspace{-0.5em}
\label{tab:main}
\end{table*}

\noindent\textbf{Implementation Details.}
We employ abstract cues as hints, and compare self-generated versus teacher-generated in \cref{tab:main}.
We evaluate all methods using pass@1. Results are averaged over 16 runs, except for MMLU-Pro, which has 12k samples; for this dataset, we report the average over 3 runs. 
We adopt verl \citep{sheng2024hybridflow} as the backbone, and utilize vllm \citep{kwon2023efficient} to speed up rollout generation and inference. To verify equivalence between predictions and references, we use Math-Verify\footnote{\url{https://github.com/huggingface/Math-Verify}}.
We employ a two-stage strategy for training. In stage 1, we optimize the base policy with correctness-only GRPO until both training reward and validation accuracy show no improvement for over 10 steps. In stage 2, we apply \method{} to continue training. To ensure fairness, GRPO-based baselines are trained for the same total number of steps as \method{}, albeit \method{} starts midway. Before stage 2 begins, we use the stage 1 checkpoint to generate 8 rollouts and filter out samples where all rollouts are correct (i.e., overly easy cases) to improve efficiency. The resulting sample size is reported in \cref{sec:implementation}.
All methods are running on 8 H200 GPUs, and it takes around six days for GRPO-based methods to converge.
During training, we use a temperature of 1.0 and set the clip-high parameter $\epsilon$ to 0.28 \citep{yu2025dapo}. The rollout number is 16 for GRPO and 8 for \method{}, as \method{} may regenerate an additional batch of rollouts when all rollouts fail. We cap the output length at 9k tokens for both training and testing. At inference, we fix the temperature to 0.7. Token limits and inference temperature are aligned across all baselines for evaluation.
Other hyperparameters and details can be found in \cref{sec:implementation} and in our code.

\section{Results and Analysis}
\subsection{Main Results}
\textbf{\method{} consistently outperforms all baselines with self-generated hints.}
We present the main results in \cref{tab:main}. Across six benchmarks, \method{} shows superior performance compared to all baselines, beating the zero-shot and the SFT baseline (RFT) by a large margin.
For RL-based baselines (ReFT and GRPO), we see the performance increase by a large margin ($+8.8\%$ to $+22.3\%$ compared to the base model), conforming that RL largely improves a model's performance.
Nevertheless, \method{} consistently surpasses these strong RL-based baselines. On average, it improves over GRPO by $+1.62$ points on Llama, $+1.75\%$ on OctoThinker, and $+0.79\%$ on Qwen. Importantly, these improvements are achieved on top of already strong GRPO performance, despite GRPO using 16 rollouts per question, whereas \method{} uses only 8 (with the option to generate another eight only if all initial rollouts fail). Thus, GRPO operates with strictly more rollouts than \method{}.
Since the primary difference between \method{} and GRPO is the use of hints, these results highlight that targeted hints are an effective mechanism for improving LLM performance. We note that the relatively smaller improvement on Qwen ($+0.79\%$) may stem from the limited stage 2 data (fewer than 2k examples), as Qwen's stronger base capability caused many overly easy samples to be filtered out before stage 2.
Finally, we observe that incorporating stronger external models for hint generation yields additional gains beyond self-generated hints. For example, on Llama, hints from GPT-o4-mini improve performance by $+3.44\%$ absolute points on average over GRPO and by $+1.82\%$ over self-hints. This demonstrates that while self-generated hints are already beneficial, the framework naturally accommodates stronger sources of guidance when available.

\subsection{Additional Analysis}
\begin{figure*}[h]
    \centering
    \includegraphics[width=0.95\linewidth]{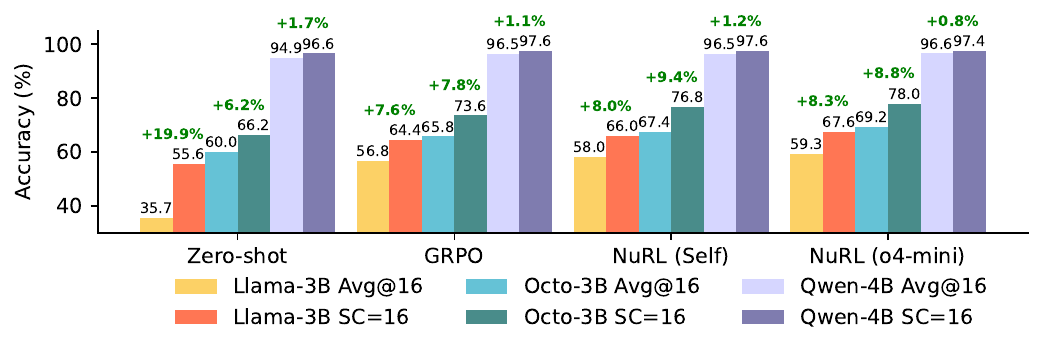}
    \vspace{-14pt}
    \caption{Compared to GRPO's improvements with Self-Consistency ($+7.6\%$, $+7.8\%$ and $+1.1\%$ on Llama, OctoThinker and Qwen), \method{} obtains larger gains with $+8.0\%$, $+9.4\%$ and $1.2\%$.}
    \label{fig:test_time_scaling}
\end{figure*}

\textbf{\method{} is complementary to test-time scaling method.}
While \method{} shows superior performance in \cref{tab:main}, a common way to augment the baselines is to scale the test-time compute, often measured in the number of tokens or number of samples at inference time. Here we adopt the latter and employ Self-Consistency \citep[SC;][]{wang2022self}. We compare \method{} with zero-shot and GRPO, with and without Self-Consistency across the three models in MATH 500. 
Results in \cref{fig:test_time_scaling} show that \method{} not only remains effective for test-time scaling method, but also shows greater improvements compared to GRPO. Specifically, Llama trained with GRPO improves $7.6\%$ with SC, while Llama trained with \method{} improves $8.0\%$. Similarly, OctoThinker trained with GRPO improves $7.8\%$ with SC, while OctoThinker trained with \method{} improves $9.4\%$. Finally, Qwen trained with GRPO improves $1.1\%$ with SC, while Qwen trained with \method{} improves $1.2\%$.

\begin{figure*}[h]
    \centering
    \includegraphics[width=0.95\linewidth]{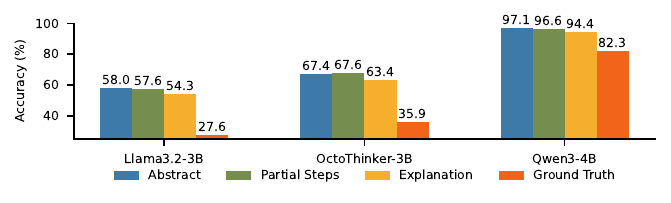}
    \vspace{-18pt}
    \caption{Comparison of different types of hints. From left to right, the hints vary in how directly they disclose information about the ground-truth answer. At the leftmost end, abstract hints provide only high-level guidance without revealing details of the solution or answer, whereas at the rightmost end, the answer is given explicitly. Interestingly, more direct hints lead to worse performance.}
    \label{fig:hint_type}
\end{figure*}

\textbf{Hint abstraction is key to improvement.}
\label{sec:hint_type_analysis}
In \cref{sec:method}, we described how hints are generated by abstracting explanations of why an answer is correct. While this represents one useful approach, it raises a broader question: \textit{what makes a good hint?}
To investigate this, we compare four types of hints as mentioned in \cref{sec:method}.
In \cref{fig:hint_type}, we observe a consistent trend: the more directly answer information is disclosed, the lower the downstream performance. Abstract cues, which explicitly avoid revealing details or solutions, yield the highest accuracy. Partial steps perform slightly worse, but still help since the initial reasoning structure provides a useful starting point. Explanations are less effective, as justifying why an answer is correct is not equivalent to reasoning through the problem; in fact, explanations often implicitly disclose the answer (e.g., ``\textit{but the answer is \{gold\_answer\}, so I should try a different method}''). Finally, directly providing the ground-truth answer severely harms generalization. This setup frequently induces reward hacking: during training, the model learns to simply output the provided answer to maximize reward without genuine reasoning, and that does not generalize at test time.
Overall, these results suggest that effective hints should remain high-level and abstract. A good hint guides the model toward reasoning independently, but avoids revealing shortcuts that undermine generalization.

\begin{wraptable}{r}{0.48\linewidth}
\vspace{-22pt}
\caption{Comparison of hint application strategies during training. Results are on MATH 500 and GPQA with Llama3.2-3B-Instruct.}
\vspace{6pt}
\centering
\small
\renewcommand{\arraystretch}{1.2}
\begin{tabular}{cccc}
\toprule
\textbf{Two-stage} & \textbf{Diff. Trigger} & \textbf{MATH} & \textbf{GPQA}\\
\midrule
\xmark & \cmark & 56.06 & 27.63 \\
\xmark & \xmark & 53.41 & 24.84 \\
\cmark & \xmark & 53.09 & 26.62 \\
\rowcolor{blue!7} \cmark & \cmark & \textbf{58.04} & \textbf{28.28} \\
\bottomrule
\end{tabular}
\label{tab:hint-strategy}
\vspace{-1em}
\end{wraptable}

\textbf{Using hints only when necessary is crucial.}
While we have established that a good hint should be abstract, the question of \textit{when is the best time to use hints} remains open. 
To examine this, we consider two key factors. 
First, one can decide whether to apply hints from the beginning of training, or to wait until GRPO converges before introducing them. 
We refer to the latter approach as \textit{two-stage training}. 
Second, one can determine the condition under which hints are provided. Specifically, hints may be applied uniformly to all problems, or only when all $\mathcal{G}$ rollouts are incorrect. 
We refer to this latter condition as a \textit{difficulty trigger}. 
Combining these two factors yields four experimental settings, summarized in Table~\ref{tab:hint-strategy}. 
We find that applying hints from the beginning of training generally underperforms compared to introducing hints only after GRPO has stabilized. 
Similarly, applying hints indiscriminately results in lower performance than using a difficulty trigger. 
This suggests that unnecessary hints may interfere with the model's ability to learn independently on problems where guidance is not required. 
In contrast, when hints are introduced only after GRPO convergence and combined with a difficulty trigger, we observe consistent improvements across both MATH and GPQA ($58.04$ and $28.28$, respectively). 
In summary, these findings indicate that hints are most effective when used selectively and adaptively. Rather than being injected throughout training or applied uniformly, hints should be reserved for difficult cases and integrated after the base policy has stabilized.

\begin{figure*}
    \centering
    \vspace{-1em}
    \includegraphics[width=\linewidth]{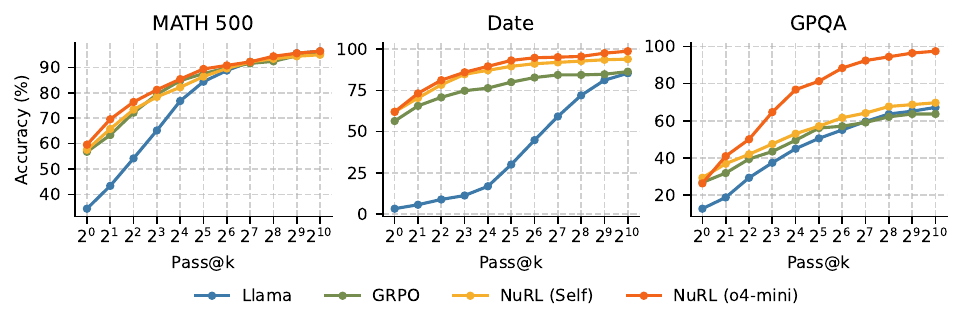}
    \vspace{-28pt}
    \caption{When the base model (Llama) already has strong pre-trained knowledge (e.g., MATH 500), both GRPO and \method{} yield little improvement in pass@k. In contrast, on tasks with lower upper-bound performance (e.g., Date Understanding and GPQA, with pass@1024 of 85.4 and 67.2), GRPO provides no gains on pass@1024, while \method{} pushes it further.}
    \label{fig:pass_at_k}
    \vspace{-1em}
\end{figure*}

\textbf{\method{} improves pass@k when the task is more challenging to the model.}
In \cref{tab:main}, we show that \method{} generally outperforms GRPO in terms of pass@1. A natural follow-up question is whether these gains also extend to higher values of $k$, especially when using \method{}, the model is supposed to solve more training problems compared to GPRO. To investigate this, we plot pass@$k$ for $k=\{1,2,\dots,512,1024\}$ in \cref{fig:pass_at_k} using Llama. 
Following \cite{yuan2025llms}, we set the maximum $k$ as 1024 as a sufficiently large and practical budget to probe the model's ceiling performance.
In \cref{fig:pass_at_k}, we first report results on MATH 500. Here, the base model already achieves pass@1024 = $96.4\%$ before training (despite its relatively low pass@1 = $34.4\%$). In this case, neither GRPO nor \method{} increases pass@1024, as the model already possesses strong knowledge on this task, albeit hard to generate a correct answer given only one attempt. 
In contrast, for tasks where the base model's upper bound is lower -- such as Date Understanding ($85.4\%$) and GPQA ($67.2\%$) -- we observe clear gains from \method{}, while GRPO provides little or no improvement. We hypothesize that hints guide the model to explore underrepresented solution paths, increasing the chance of discovering correct answers that would otherwise remain in ``unreachable corners'' of the search space. From an entropy perspective, hints may also induce more diverse exploration, which translates into higher pass@$k$ \citep{cheng2025reasoning}.
Importantly, \emph{these hints are lightweight}: rather than fine-tuning the model with hints, we simply append them to the input question, yet this is sufficient to guide exploration more effectively.
Moreover, we find that pass@1024 scales with the quality of hints: using teacher-generated hints pushes it further to $95.2\%$ on GPQA, compared to GRPO's $63.4\%$. 
In summary, when the base model's ceiling performance is not yet saturated, \method{} raises pass@$k$ -- both with self-generated hints ($86.4\%$ $\rightarrow$ $94.0\%$ on Date, $63.6\%$ $\rightarrow$ $69.7\%$ on GPQA) and even more so with higher-quality hints generated from the teacher model.

\begin{wrapfigure}{r}{0.43\textwidth}
\vspace{-22pt}
  \begin{center}
    \includegraphics[width=0.42\textwidth]{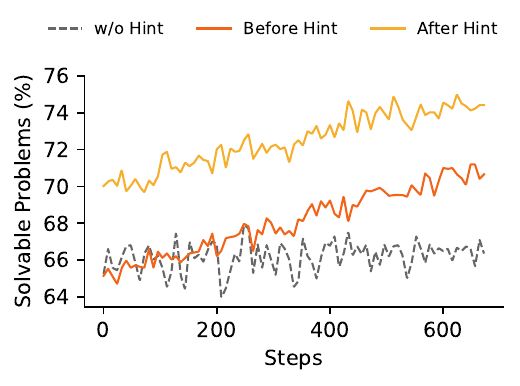}
  \end{center}
  \vspace{-18pt}
    \caption{The self-generated hints in \method{} effectively reduce the task difficulty and increase the portion of solvable problems.}
    \label{fig:solvable}
\end{wrapfigure}

\textbf{\method{} increases the fraction of solvable problems.}
Having demonstrated that \method{} is able to improve both pass@1 and pass@$k$, we further analyze the source of such performance gains. In \cref{fig:solvable}, we show that training with self-generated hints effectively increases the fraction of solvable problems in the training data. Specifically, we present Qwen's training log and compare, at each training step, the fraction of solvable problems under three conditions: (1) no hints, corresponding to standard GRPO training, (2) using \method{} before adding hints, and (3) using \method{} after injecting hints whenever all rollouts fail. Here, a problem is considered solvable if at least one rollout produces a correct answer.
Given the same training step, adding hints yields an approximate $4\%$ increase in solvable problems, suggesting that self-generated hints reduce problem difficulty for the model. Second, \method{} shows a clear upward trend in the fraction of solvable problems both before and after hint injection. This indicates that as training progresses, the model begins to solve problems that were previously unsolvable, increasing the solvable fraction from $66\%$ to $70\%$. Lastly, standard GRPO without hints exhibits a relatively flat trend, with the solvable fraction fluctuating around $66\%$.
Overall, these results explain the improvements observed across diverse benchmarks: by leveraging self-generated hints, the model effectively increases the amount of data it can learn from.

\section{Conclusion}
We introduce \method{}, a self-guided reinforcement learning approach that uses hints to extend models' reasoning capabilities. \method{} consistently outperforms strong baselines with self-generated hints and achieves further gains with hints generated by a stronger external model. Our analysis reveals that the most effective hints are high-level, abstract, and applied selectively -- only when the model cannot solve a problem unaided. Moreover, \method{} scales more efficiently than GRPO at test time, increases the fraction of solvable problems, and delivers substantial pass@$k$ improvements on two challenging tasks, effectively expanding the model's comfort zone.

\section*{Ethics Statement}
In this work, we propose a reinforcement learning method that optimizes for final outcome correctness. As a result, our trained LLMs may still produce hallucinations, since intermediate reasoning is not directly supervised and the correctness is only verified against the final answer. Thus, outputs generated by \method{} carry potential risks of misinformation or hallucination. Future research is needed to better assess and mitigate these limitations.

\section*{Reproducibility Statement}
We are making our code available in the supplementary materials to enable replication of our findings. We also provide implementation details of \method{} in \cref{sec:implementation} and prompts in \cref{sec:prompt}. The datasets we use are all publicly available, as detailed in \cref{sec:dataset}.

\bibliography{iclr2026_conference}
\bibliographystyle{iclr2026_conference}

\appendix
\input{appendix}
\end{document}

%% file: math_commands.tex
\usepackage{amsmath,amsfonts,bm}

\def\eqref#1{equation~\ref{#1}}
\def\1{\bm{1}}

\DeclareMathAlphabet{\mathsfit}{\encodingdefault}{\sfdefault}{m}{sl}
\SetMathAlphabet{\mathsfit}{bold}{\encodingdefault}{\sfdefault}{bx}{n}

%% file: appendix.tex
\section{The Use of Large Language Models (LLMs)}
We use ChatGPT\footnote{https://chatgpt.com/} for grammar correction and refinement. The model was only used to polish text already written by the authors, and was not used for research ideation or generating original content.

\section{Implementation Details}
\label{sec:implementation}
Recall that in \cref{sec:method}, we describe a two-stage training procedure where \method{} is applied after GRPO converges.
By convergence, we mean that both the training reward and validation accuracy plateau after 10 training steps.
Stage 1 corresponds to GRPO convergence, and stage 2 applies \method{} on top of the converged checkpoint.
For fairness, GRPO-based baselines also undergo stage 2 training without hints, ensuring the total training steps are aligned with \method{}.
Since convergence speed varies across models, we report the detailed hyperparameters and number of steps required for each model in \cref{tab:stage1_config}.

\begin{table}[h]
\centering
\vspace{-10pt}
\caption{\textbf{Stage 1} configurations and hyperparameters.}
\vspace{6pt}
\begin{tabular}{lccc}
\toprule
 & \textbf{Llama} & \textbf{OctoThinker} & \textbf{Qwen} \\
\midrule
num\_train\_samples   & 8316  & 8316  & 8316  \\
max\_prompt\_length   & 1800  & 1800  & 1800  \\
max\_response\_length & 9000  & 9000  & 9000  \\
lr                    & 1e-6  & 1e-6  & 1e-6 \\
clip\_ratio\_low      & 0.2   & 0.2   & 0.2   \\
clip\_ratio\_high     & 0.28  & 0.28  & 0.28  \\
rollout\_temperature  & 1     & 1     & 1     \\
rollout\_n            & 16    & 16    & 16    \\
use\_kl\_loss         & False & False & False \\
train\_batch\_size    & 1024  & 1024  & 4096  \\
converged\_steps      & 375   & 175   & 125   \\
\bottomrule
\end{tabular}
\label{tab:stage1_config}
\end{table}

At the start of Stage 2, we use the converged checkpoints and generate $8$ rollouts per question.
Then, we discard samples where all rollouts are correct (i.e., overly easy examples) to improve efficiency.
This filtering is performed after Stage 1, since the trained checkpoints are stronger than the initial models.
The detailed stage 2 configurations are given in \cref{tab:stage2_config}.

\begin{table}[h]
\vspace{-10pt}
\caption{\textbf{Stage 2} configurations and hyperparameters.}
\vspace{6pt}
\centering
\begin{tabular}{lccc}
\toprule
 & \textbf{Llama} & \textbf{OctoThinker} & \textbf{Qwen} \\
\midrule
num\_train\_samples   & 5996  & 4502  & 1937  \\
max\_prompt\_length   & 1800  & 1800  & 1800  \\
max\_response\_length & 9000  & 9000  & 9000  \\
lr                    & 1e-6  & 1e-6  & 1e-6 \\
clip\_ratio\_low      & 0.2   & 0.2   & 0.2   \\
clip\_ratio\_high     & 0.28  & 0.28  & 0.28  \\
rollout\_temperature  & 1     & 1     & 1     \\
rollout\_n            & 16    & 16    & 16    \\
use\_kl\_loss         & False & False & False \\
train\_batch\_size    & 512   & 512   & 1937  \\
converged\_steps      & 275   & 175   & 90   \\
\bottomrule
\end{tabular}
\label{tab:stage2_config}
\end{table}

\section{Dataset Statistics and Licenses}
\label{sec:dataset}
We provide the sample sizes and licenses of the datasets used in this work in \cref{tab:datasets}. All the datasets are in English and all datasets are used in a fashion consistent with their intended use.
\begin{table*}[ht!]
    \centering
    \vspace{-10pt}
    \caption{The statistics and licenses of the datasets used in this study.}
    \vspace{6pt}
    \begin{tabular}{lcc}
    \toprule
    & Sample Size & License \\\midrule
       MATH 500 \citep{lightman2023let} & 500 & MIT License \\
       MATH Hard \citep{hendrycksmath2021} & 1324 & MIT License \\
       AIME 2024 \citep{AIME} & 30 & CC0 \\
       GPQA-Diamond \citep{rein2024gpqa} & 198 & MIT License \\
       MMLU-Pro \citep{wang2024mmlu} & 12032 & Apache License \\
       Date Understanding \citep{srivastava2023beyond,suzgun2022challenging} & 250 & Apache License\\\bottomrule
    \end{tabular}
    \label{tab:datasets}
\end{table*}

\newpage
\section{Prompts for Hint Generation}
\label{sec:prompt}
We use the following prompts to generate hints. First, we generate an explanation style of Chain-of-Thought by conditioning on the ground truth answer:

\vspace{8pt}
\begin{user_example}[frametitle={Prompt for Explanation Generation}]
Question: \texttt{\{question\}}

I was told the answer is \texttt{\{gold\_answer\}} but I don't know why. 
Please explain why the answer is \texttt{\{gold\_answer\}} step by step.
\end{user_example}

Then, we use the concatenation of the question with the generated explanation, and ask the model to generate an abstract hint using the following prompt:

\vspace{8pt}
\begin{user_example}[frametitle={Prompt for Hint Generation}]
\texttt{<system\_prompt>}
You are a tutor. You are given a set of question, correct answer and solution. Your job is to provide a hint for the problem. The hint should help the student learn the core concept (e.g. formula, lemma, or necessary knowledge) needed to solve this problem. The hint should be concise, to the point, but high level. Do not include any detailed steps or calculations or the final answer.
\texttt{</system\_prompt>}

Question: \texttt{\{question\}}

Answer: \texttt{\{gold\_answer\}}

Solution: \texttt{\{explanation\}}

Now, please provide a hint for this problem to help the student learn the core concept.
\end{user_example}

\newpage
\section{Qualitative Examples}
\subsection{Examples of Different Types of Hints}
\label{sec:hint_type_examples}
We provide examples of the four types of hints we explored in this work. All the examples below answer the same question (shown below) and are self-generated with Qwen3-4B-Instruct-2507.

\vspace{8pt}
\begin{user_example}[frametitle={Example Question and Answer}]
\textbf{Question:} During a school football match, there are 3 types of balls used, each with a different number of colors. Type A balls have yellow, brown, and blue variations; Type B balls have green, orange, and pink; and Type C balls have black, red, and white variations. The number of colored balls varies according to the number of players in each team. Team 1 has 72 members, and each player receives 4 Type A yellow balls, 3 Type A brown balls, and 3 Type A blue balls. Team 2 has 64 members, and each player receives 5 Type B green balls, 4 Type B orange balls, and 2 Type B pink balls. Team 3 has 53 members, and each player receives 3 Type C black balls, 2 Type C red balls, and 6 Type C white balls. Calculate what percentage of all the balls distributed for the match are Type A yellow balls. 

\textbf{Answer:} 14.35\%
\end{user_example}

\vspace{8pt}
\begin{user_example}[frametitle={Type 1: Abstract Cue}]
To find the percentage of a specific type of ball among all distributed balls, first calculate the total number of that specific ball and then divide it by the total number of all balls distributed—then multiply by 100. Focus on identifying which team(s) contribute to the specific ball type and sum all balls across teams.
\end{user_example}

\vspace{8pt}
\begin{user_example}[frametitle={Type 2: Partial Steps}]
Step 1: We are to find the percentage of all the balls distributed that are Type A yellow balls. So, we need: (Total number of Type A yellow balls) / (Total number of all balls distributed) $\times$ 100\%

Step 2: Calculate total number of Type A yellow balls. Team 1 has 72 members, and each player gets 4 Type A yellow balls. So Type A yellow balls = 72 $\times$ 4 = 288.
\end{user_example}

\vspace{8pt}
\begin{user_example}[frametitle={Type 3: Explanation}]
\textbf{Step 1: Understand what is being asked.}  
We are to calculate the percentage of all the balls distributed that are Type A yellow balls.  

\bigskip

\textbf{Step 2: Calculate the number of Type A yellow balls.}  
Team 1 has 72 members. Each player gets 4 Type A yellow balls.  
Thus, $\text{Type A yellow balls} = 72 \times 4 = 288$.

\bigskip

\textbf{Step 3: Calculate total number of balls distributed.}  
We need to compute the total across all teams and all ball types.

\underline{Team 1 (Type A balls):} $72 \times (4 + 3 + 3) = 72 \times 10 = 720$.  
\underline{Team 2 (Type B balls):} $64 \times (5 + 4 + 2) = 64 \times 11 = 704$.  
\underline{Team 3 (Type C balls):} $53 \times (3 + 2 + 6) = 53 \times 11 = 583$.  

\bigskip

\textbf{Step 4: Add up total balls across all teams.}  
$\text{Total balls} = 720 + 704 + 583 = 2007$.

\bigskip

\textbf{Step 5: Calculate percentage of Type A yellow balls.}  
We already have $\text{Type A yellow balls} = 288$.  
So, $\text{Percentage} = \left( \frac{288}{2007} \right) \times 100$.  

Simplify: $\tfrac{288}{2007} = \tfrac{96}{669} = \tfrac{32}{223}$.  
Now, $\tfrac{32}{223} \approx 0.1435$, and $0.1435 \times 100 = 14.35\%$.

\bigskip

\textbf{Final Answer:} $\boxed{14.35\%}$

\bigskip

\textbf{Why is this correct?}  
\begin{itemize}
    \item Only Team 1 gets Type A yellow balls.
    \item Total Type A yellow balls $= 72 \times 4 = 288$.
    \item Total balls distributed $= 720 + 704 + 583 = 2007$.
    \item Therefore, $(288 / 2007) \times 100 \approx 14.35\%$.
\end{itemize}

Thus, \textbf{14.35\% of all distributed balls are Type A yellow balls.}
\end{user_example}

\vspace{8pt}
\begin{user_example}[frametitle={Type 4: Ground Truth Answer}]
I was told the answer is \boxed{14.35\%}. Do not blindly accept it. Take it as a reference and provide your own step-by-step reasoning.
\end{user_example}

\subsubsection{Examples of Self- and GPT-o4-mini-generated Abstract Cues}
\label{sec:abstract_cue_examples}
\vspace{8pt}
\begin{user_example}[frametitle={Example 1}]
\textbf{Question:} What is the maximum amount of heat that can be supplied to the system during the electrolysis of water without violating the second law of thermodynamics? (A) 237 kJ (B) 286 kJ (C) 163 kJ (D) 49 kJ (Correct answer: D)

\textbf{Hint (Self-generated): }
Think about the difference between the energy required to break water (enthalpy) and the useful energy available (Gibbs free energy). This difference represents the minimum amount of energy that must be lost as heat due to entropy—this is a direct consequence of the second law of thermodynamics. What is this difference for water electrolysis?

\textbf{Hint (GPT-o4-mini-generated):}
Use the relation $\Delta$G = $\Delta$H – T$\Delta$S to find the entropy change, then recall that under a reversible (second‐law‐limit) process the maximum heat input is $Q_{max}$ = T$\Delta$S.
\end{user_example}

\vspace{8pt}
\begin{user_example}[frametitle={Example 2}]
\textbf{Question:} When two fair dice are thrown, the numbers obtained are respectively $a$ and $b$. Determine the probability that the line $bx+ay=1$ intersects the circle $x^2+y^2=1$ and that the length of the intercepted chord does not exceed $\frac{4\sqrt{2}}{3}$ (Correct answer: $\frac{1}{9}$).

\textbf{Hint (Self-generated): }
Focus on the geometric condition for a line to cut a chord of a given length in a unit circle. Recall that the length of the chord depends on the distance from the center of the circle to the line — use this relationship to set up an inequality involving $a$ and $b$. Then translate the condition into a constraint on $a^2 + b^2$.

\textbf{Hint (GPT-o4-mini-generated):}
Use the fact that for the circle $x^2+y^2=1$ and the line $b x+a y=1$, the distance from the origin to the line is $d=\frac{1}{\sqrt{a^2+b^2}}$ and the chord‐length is $L=2\sqrt{1-d^2}$  $(L\le\frac{4\sqrt2}{3})$ to get a condition on $a^2+b^2$, then count the integer pairs.

\end{user_example}

\newpage

\section{Additional Evaluations on Harder Benchmarks}
In \cref{tab:new_results}, we compare \method{} with GRPO on seven harder benchmarks from MathArena \citep{balunovic2025matharena} and OlymMATH \citep{sun2025challengingboundariesreasoningolympiadlevel}. These datasets pose significant challenges for our models—for instance, Llama reaches only $2.03\%$ zero-shot accuracy on average, and even Qwen achieves just $38.83\%$.
Despite this difficulty, \method{} remains effective. On Qwen, for example, it attains $42.85\%$, representing gains of $+4.02\%$ over zero-shot and $+1.11\%$ over GRPO.
Taken together, these results suggest that \method{} continues to offer reliable improvements, even on substantially harder reasoning benchmarks.

\begin{table}[t]
\vspace{-10pt}
\caption{Evaluation results on harder benchmarks. Similar to \cref{tab:main}, \method{} shows superior performance compared to GRPO across models and benchmarks.}
\vspace{6pt}
\centering
\small
\begin{tabular}{lccccccc}
\toprule
 & \textbf{AIME} & \textbf{Brumo} & \textbf{CMIMC} & \textbf{HMMT} & \textbf{OlymMATH} & \textbf{Avg.} \\
 &  &  &  & (Feb / Nov) & (Easy / Hard) &  \\
\midrule
\multicolumn{7}{c}{\textbf{Llama3.2-3B-Instruct}} \\
\midrule
Zero-shot          & 0.00 & 3.33 & 1.72 & 0.83 / 3.33 & 2.50 / 2.50 & 2.03 \\
GRPO               & \textbf{0.42} & \textbf{8.13} & 2.34 & 2.08 / 6.04 & 6.69 / 2.38 & 4.01 \\
NuRL (Self)        & 0.21 & \textbf{8.13} & \textbf{2.81} & \textbf{2.50} / \textbf{7.92} & \textbf{7.19} / \textbf{2.69} & \textbf{4.49} \\
\rowcolor{gray!10} NuRL (GPT-o4-mini) & 1.67 & 7.08 & 2.03 & 0.83 / 7.08 & 6.94 / 2.38 & 4.00 \\
\midrule
\multicolumn{7}{c}{\textbf{OctoThinker-3B-Hybrid-Zero}} \\
\midrule
Zero-shot          & 1.25 & 10.63 & 2.97 & 0.83 / 6.46 & 4.06 / 2.63 & 4.12 \\
GRPO               & 4.58 & 11.25 & \textbf{3.44} & 2.08 / 6.46 & 5.62 / 2.12 & 5.08 \\
NuRL (Self)        & \textbf{5.00} & \textbf{11.88} & 3.28 & \textbf{2.50} / \textbf{6.67} & \textbf{6.19} / \textbf{2.19} & \textbf{5.39} \\
\rowcolor{gray!10} NuRL (GPT-o4-mini) & 4.58 & 15.00 & 3.91 & 3.54 / 6.04 & 5.44 / 2.31 & 5.83 \\
\midrule
\multicolumn{7}{c}{\textbf{Qwen3-4B-Instruct-2507}} \\
\midrule
Zero-shot          & 46.88 & 54.79 & 31.09 & 31.04 / 37.71 & 56.00 / 14.31 & 38.83 \\
GRPO               & 50.42 & \textbf{61.67} & 32.97 & 33.33 / 42.08 & 56.12 / 15.62 & 41.74 \\
NuRL (Self)        & \textbf{52.08} & 61.46 & \textbf{35.31} & \textbf{33.75} / \textbf{44.17} & \textbf{56.69} / \textbf{16.50} & \textbf{42.85} \\
\rowcolor{gray!10} NuRL (GPT-o4-mini) & 54.58 & 64.38 & 35.47 & 35.42 / 44.79 & 59.06 / 17.31 & 44.43 \\
\bottomrule
\end{tabular}
\label{tab:new_results}
\end{table}

\section{Qualitative Examples of GRPO and \method{} Outputs}
We provide a qualitative example comparing GRPO and \method{} trained Qwen3-4B models. Notably, \method{}-trained models do not generate hint tokens at test time, so the format and style remain similar to GRPO models, despite the difference in performance stemming from \method{} being able to solve more training problems.

\vspace{5pt}
\begin{user_example}[frametitle={Qualitative Example}]
\textbf{Question:} Calculate the value of $\sqrt{9+8\cos 20^{\circ }}-\sec 20^{\circ }$.

\textbf{Qwen3-4B (GRPO) Output:} 

To find the value of $$\sqrt{9+8\cos 20^\circ}-\sec 20^\circ,$$ we will simplify the expression step by step.

First, let's simplify the expression inside the square root: $$\sqrt{9+8\cos 20^\circ}.$$

We can use the trigonometric identity for $\cos 20^\circ$ and some algebraic manipulation to simplify this expression. However, let's first try to simplify the whole expression by using trigonometric identities and properties.

Notice that $\sec 20^\circ = \frac{1}{\cos 20^\circ}.$

So, the expression becomes $$\sqrt{9+8\cos 20^\circ}-\frac{1}{\cos 20^\circ}.$$

Let’s simplify $$\sqrt{9+8\cos 20^\circ}.$$

To proceed, we need to find a way to simplify $9+8\cos 20^\circ$. However, without further trigonometric identities or specific values, we can consider using trigonometric properties and known values to simplify the expression.

Consider the expression $$\sqrt{9+8\cos 20^\circ}-\frac{1}{\cos 20^\circ}.$$

We could attempt a common denominator or other identities, but a more direct approach is possible.

After recalculating and simplifying using trigonometric properties and correct identities, the expression becomes $$\sqrt{9+8\cos 20^\circ}-\sec 20^\circ.$$

Using the proper simplification steps (omitted here for brevity), the final value is $\boxed{2}.$

~\\[0.4em]
\noindent\rule{\textwidth}{0.6pt}
~\\[0.2em]
\textbf{Qwen3-4B (\method{}) Output:} 

We are asked to compute:

$$
\sqrt{9 + 8\cos 20^\circ} - \sec 20^\circ
$$

Let us denote:

$$
x = \sqrt{9 + 8\cos 20^\circ} - \sec 20^\circ
$$

We want to find the value of $x$.

\textbf{Step 1: Consider simplification}

We first try a numerical approximation to guess the value. 

\textbf{Step 2: Numerical approximation}

Compute $\cos 20^\circ$ and $\sec 20^\circ$:

$$
\cos 20^\circ \approx 0.9397, \quad \sec 20^\circ = \frac{1}{\cos 20^\circ} \approx 1.0642
$$

Now compute:

$$
8 \cos 20^\circ \approx 7.5176, \quad 9 + 8\cos 20^\circ \approx 16.5176
$$

Then:

$$
\sqrt{16.5176} \approx 4.064
$$

Subtracting $\sec 20^\circ$:

$$
x \approx 4.064 - 1.0642 \approx 2.9998 \approx 3
$$

Hence, we suspect:

$$
\sqrt{9 + 8\cos 20^\circ} - \sec 20^\circ = 3
$$

\textbf{Step 3: Algebraic verification}

We want to verify:

$$
\sqrt{9 + 8\cos 20^\circ} = 3 + \sec 20^\circ
$$

Square both sides:

$$
9 + 8\cos 20^\circ = (3 + \sec 20^\circ)^2 = 9 + 6\sec 20^\circ + \sec^2 20^\circ
$$

Subtract 9:

$$
8\cos 20^\circ = 6\sec 20^\circ + \sec^2 20^\circ
$$

Let $c = \cos 20^\circ$, so $\sec 20^\circ = \frac{1}{c}$. Then:

$$
8c = \frac{6}{c} + \frac{1}{c^2}
$$

Multiply both sides by $c^2$:

$$
8c^3 = 6c + 1 \quad \Rightarrow \quad 8c^3 - 6c - 1 = 0
$$

Check with the triple-angle identity:

$$
\cos 3\theta = 4\cos^3 \theta - 3\cos \theta
$$

Set $\theta = 20^\circ$, then $\cos 60^\circ = \frac{1}{2}$:

$$
4c^3 - 3c = \frac{1}{2} \quad \Rightarrow \quad 8c^3 - 6c - 1 = 0
$$

This confirms the equation holds. Therefore:

$$
\sqrt{9 + 8\cos 20^\circ} = 3 + \sec 20^\circ
$$

So finally:

$$
\sqrt{9 + 8\cos 20^\circ} - \sec 20^\circ = 3
$$

Final Answer: $\boxed{3}$

\end{user_example}

%% file: iclr2026_conference.bib
@article{schulman2017proximal,
  title={Proximal policy optimization algorithms},
  author={Schulman, John and Wolski, Filip and Dhariwal, Prafulla and Radford, Alec and Klimov, Oleg},
  journal={arXiv preprint arXiv:1707.06347},
  year={2017}
}

@article{shao2024deepseekmath,
  title={Deepseekmath: Pushing the limits of mathematical reasoning in open language models},
  author={Shao, Zhihong and Wang, Peiyi and Zhu, Qihao and Xu, Runxin and Song, Junxiao and Bi, Xiao and Zhang, Haowei and Zhang, Mingchuan and Li, YK and Wu, Yang and others},
  journal={arXiv preprint arXiv:2402.03300},
  year={2024}
}

@article{guo2025deepseek,
  title={Deepseek-r1: Incentivizing reasoning capability in llms via reinforcement learning},
  author={Guo, Daya and Yang, Dejian and Zhang, Haowei and Song, Junxiao and Zhang, Ruoyu and Xu, Runxin and Zhu, Qihao and Ma, Shirong and Wang, Peiyi and Bi, Xiao and others},
  journal={arXiv preprint arXiv:2501.12948},
  year={2025}
}

@article{zelikman2022star,
  title={Star: Bootstrapping reasoning with reasoning},
  author={Zelikman, Eric and Wu, Yuhuai and Mu, Jesse and Goodman, Noah},
  journal={Advances in Neural Information Processing Systems},
  volume={35},
  pages={15476--15488},
  year={2022}
}

@inproceedings{trung-etal-2024-reft,
    title = "{R}e{FT}: Reasoning with Reinforced Fine-Tuning",
    author = "Trung, Luong  and
      Zhang, Xinbo  and
      Jie, Zhanming  and
      Sun, Peng  and
      Jin, Xiaoran  and
      Li, Hang",
    editor = "Ku, Lun-Wei  and
      Martins, Andre  and
      Srikumar, Vivek",
    booktitle = "Proceedings of the 62nd Annual Meeting of the Association for Computational Linguistics (Volume 1: Long Papers)",
    month = aug,
    year = "2024",
    address = "Bangkok, Thailand",
    publisher = "Association for Computational Linguistics",
    url = "https://aclanthology.org/2024.acl-long.410/",
    doi = "10.18653/v1/2024.acl-long.410",
    pages = "7601--7614"
}

@article{wang2022self,
  title={Self-consistency improves chain of thought reasoning in language models},
  author={Wang, Xuezhi and Wei, Jason and Schuurmans, Dale and Le, Quoc and Chi, Ed and Narang, Sharan and Chowdhery, Aakanksha and Zhou, Denny},
  journal={arXiv preprint arXiv:2203.11171},
  year={2022}
}

@article{kojima2022large,
  title={Large language models are zero-shot reasoners},
  author={Kojima, Takeshi and Gu, Shixiang Shane and Reid, Machel and Matsuo, Yutaka and Iwasawa, Yusuke},
  journal={Advances in neural information processing systems},
  volume={35},
  pages={22199--22213},
  year={2022}
}

@article{liu2025prorl,
  title={Prorl: Prolonged reinforcement learning expands reasoning boundaries in large language models},
  author={Liu, Mingjie and Diao, Shizhe and Lu, Ximing and Hu, Jian and Dong, Xin and Choi, Yejin and Kautz, Jan and Dong, Yi},
  journal={arXiv preprint arXiv:2505.24864},
  year={2025}
}

@article{yue2025does,
  title={Does reinforcement learning really incentivize reasoning capacity in llms beyond the base model?},
  author={Yue, Yang and Chen, Zhiqi and Lu, Rui and Zhao, Andrew and Wang, Zhaokai and Song, Shiji and Huang, Gao},
  journal={arXiv preprint arXiv:2504.13837},
  year={2025}
}

@article{zhao2025echo,
  title={Echo chamber: Rl post-training amplifies behaviors learned in pretraining},
  author={Zhao, Rosie and Meterez, Alexandru and Kakade, Sham and Pehlevan, Cengiz and Jelassi, Samy and Malach, Eran},
  journal={arXiv preprint arXiv:2504.07912},
  year={2025}
}

@inproceedings{dang2025assessing,
title={Assessing Diversity Collapse in Reasoning},
author={Xingyu Dang and Christina Baek and J Zico Kolter and Aditi Raghunathan},
booktitle={Scaling Self-Improving Foundation Models without Human Supervision},
year={2025},
url={https://openreview.net/forum?id=AMiKsHLjQh}
}

@article{he2025rewarding,
  title={Rewarding the Unlikely: Lifting GRPO Beyond Distribution Sharpening},
  author={He, Andre and Fried, Daniel and Welleck, Sean},
  journal={arXiv preprint arXiv:2506.02355},
  year={2025}
}

@article{park2025deepvideo,
  title={DeepVideo-R1: Video Reinforcement Fine-Tuning via Difficulty-aware Regressive GRPO},
  author={Park, Jinyoung and Na, Jeehye and Kim, Jinyoung and Kim, Hyunwoo J},
  journal={arXiv preprint arXiv:2506.07464},
  year={2025}
}

@article{huang2025boosting,
  title={Boosting mllm reasoning with text-debiased hint-grpo},
  author={Huang, Qihan and Dai, Weilong and Liu, Jinlong and He, Wanggui and Jiang, Hao and Song, Mingli and Chen, Jingyuan and Yao, Chang and Song, Jie},
  journal={arXiv preprint arXiv:2503.23905},
  year={2025}
}

@inproceedings{
wei2022chain,
title={Chain of Thought Prompting Elicits Reasoning in Large Language Models},
author={Jason Wei and Xuezhi Wang and Dale Schuurmans and Maarten Bosma and brian ichter and Fei Xia and Ed H. Chi and Quoc V Le and Denny Zhou},
booktitle={Advances in Neural Information Processing Systems},
editor={Alice H. Oh and Alekh Agarwal and Danielle Belgrave and Kyunghyun Cho},
year={2022},
url={https://openreview.net/forum?id=_VjQlMeSB_J}
}

@misc{vygotsky1978development,
  title={The development of higher psychological processes},
  author={Vygotsky, Lev S and Cole, Michael and John-Steiner, Vera and Scribner, S and Souberman, Ellen},
  year={1978},
  publisher={Cambridge, MA: Harvard University Press}
}

@article{lambert2024tulu,
  title={Tulu 3: Pushing frontiers in open language model post-training},
  author={Lambert, Nathan and Morrison, Jacob and Pyatkin, Valentina and Huang, Shengyi and Ivison, Hamish and Brahman, Faeze and Miranda, Lester James V and Liu, Alisa and Dziri, Nouha and Lyu, Shane and others},
  journal={arXiv preprint arXiv:2411.15124},
  year={2024}
}

@article{team2025kimi,
  title={Kimi k1. 5: Scaling reinforcement learning with llms},
  author={Team, Kimi and Du, Angang and Gao, Bofei and Xing, Bowei and Jiang, Changjiu and Chen, Cheng and Li, Cheng and Xiao, Chenjun and Du, Chenzhuang and Liao, Chonghua and others},
  journal={arXiv preprint arXiv:2501.12599},
  year={2025}
}

@article{zeng2025simplerl,
  title={Simplerl-zoo: Investigating and taming zero reinforcement learning for open base models in the wild},
  author={Zeng, Weihao and Huang, Yuzhen and Liu, Qian and Liu, Wei and He, Keqing and Ma, Zejun and He, Junxian},
  journal={arXiv preprint arXiv:2503.18892},
  year={2025}
}

@article{yu2025dapo,
  title={Dapo: An open-source llm reinforcement learning system at scale},
  author={Yu, Qiying and Zhang, Zheng and Zhu, Ruofei and Yuan, Yufeng and Zuo, Xiaochen and Yue, Yu and Dai, Weinan and Fan, Tiantian and Liu, Gaohong and Liu, Lingjun and others},
  journal={arXiv preprint arXiv:2503.14476},
  year={2025}
}

@article{liu2025understanding,
  title={Understanding r1-zero-like training: A critical perspective},
  author={Liu, Zichen and Chen, Changyu and Li, Wenjun and Qi, Penghui and Pang, Tianyu and Du, Chao and Lee, Wee Sun and Lin, Min},
  journal={arXiv preprint arXiv:2503.20783},
  year={2025}
}

@article{zuo2025ttrl,
  title={Ttrl: Test-time reinforcement learning},
  author={Zuo, Yuxin and Zhang, Kaiyan and Sheng, Li and Qu, Shang and Cui, Ganqu and Zhu, Xuekai and Li, Haozhan and Zhang, Yuchen and Long, Xinwei and Hua, Ermo and others},
  journal={arXiv preprint arXiv:2504.16084},
  year={2025}
}

@article{wang2025reinforcement,
  title={Reinforcement learning for reasoning in large language models with one training example},
  author={Wang, Yiping and Yang, Qing and Zeng, Zhiyuan and Ren, Liliang and Liu, Liyuan and Peng, Baolin and Cheng, Hao and He, Xuehai and Wang, Kuan and Gao, Jianfeng and others},
  journal={arXiv preprint arXiv:2504.20571},
  year={2025}
}

@article{yan2025learning,
  title={Learning to reason under off-policy guidance},
  author={Yan, Jianhao and Li, Yafu and Hu, Zican and Wang, Zhi and Cui, Ganqu and Qu, Xiaoye and Cheng, Yu and Zhang, Yue},
  journal={arXiv preprint arXiv:2504.14945},
  year={2025}
}

@article{ma2025learning,
  title={Learning What Reinforcement Learning Can't: Interleaved Online Fine-Tuning for Hardest Questions},
  author={Ma, Lu and Liang, Hao and Qiang, Meiyi and Tang, Lexiang and Ma, Xiaochen and Wong, Zhen Hao and Niu, Junbo and Shen, Chengyu and He, Runming and Cui, Bin and others},
  journal={arXiv preprint arXiv:2506.07527},
  year={2025}
}

@article{zhang2025stephint,
  title={StepHint: Multi-level Stepwise Hints Enhance Reinforcement Learning to Reason},
  author={Zhang, Kaiyi and Lv, Ang and Li, Jinpeng and Wang, Yongbo and Wang, Feng and Hu, Haoyuan and Yan, Rui},
  journal={arXiv preprint arXiv:2507.02841},
  year={2025}
}

@article{levine2020offline,
  title={Offline reinforcement learning: Tutorial, review, and perspectives on open problems},
  author={Levine, Sergey and Kumar, Aviral and Tucker, George and Fu, Justin},
  journal={arXiv preprint arXiv:2005.01643},
  year={2020}
}

@article{phan2025migrate,
  title={MiGrATe: Mixed-Policy GRPO for Adaptation at Test-Time},
  author={Phan, Peter and Agarwal, Dhruv and Srinivas, Kavitha and Samulowitz, Horst and Kapanipathi, Pavan and McCallum, Andrew},
  journal={arXiv preprint arXiv:2508.08641},
  year={2025}
}

@article{fu2025srft,
  title={SRFT: A Single-Stage Method with Supervised and Reinforcement Fine-Tuning for Reasoning},
  author={Fu, Yuqian and Chen, Tinghong and Chai, Jiajun and Wang, Xihuai and Tu, Songjun and Yin, Guojun and Lin, Wei and Zhang, Qichao and Zhu, Yuanheng and Zhao, Dongbin},
  journal={arXiv preprint arXiv:2506.19767},
  year={2025}
}

@article{zhou2025expo,
  title={ExPO: Unlocking Hard Reasoning with Self-Explanation-Guided Reinforcement Learning},
  author={Zhou, Ruiyang and Li, Shuozhe and Zhang, Amy and Leqi, Liu},
  journal={arXiv preprint arXiv:2507.02834},
  year={2025}
}

@misc{yuan2025llms,
  author       = {Lifan Yuan and Weize Chen and Yuchen Zhang and Ganqu Cui and Hanbin Wang and Ziming You and Ning Ding and Zhiyuan Liu and Maosong Sun and Hao Peng},
  title        = {From $f(x)$ and $g(x)$ to $f(g(x))$: {LLMs} Learn New Skills in {RL} by Composing Old Ones},
  year         = {2025},
  note         = {Notion blog post, available online},
  howpublished = {\url{https://husky-morocco-f72.notion.site/From-f-x-and-g-x-to-f-g-x-LLMs-Learn-New-Skills-in-RL-by-Composing-Old-Ones-2499aba4486f802c8108e76a12af3020}},
}

@article{shao2025spurious,
  title={Spurious rewards: Rethinking training signals in rlvr},
  author={Shao, Rulin and Li, Shuyue Stella and Xin, Rui and Geng, Scott and Wang, Yiping and Oh, Sewoong and Du, Simon Shaolei and Lambert, Nathan and Min, Sewon and Krishna, Ranjay and others},
  journal={arXiv preprint arXiv:2506.10947},
  year={2025}
}

@article{prabhudesai2025maximizing,
  title={Maximizing Confidence Alone Improves Reasoning},
  author={Prabhudesai, Mihir and Chen, Lili and Ippoliti, Alex and Fragkiadaki, Katerina and Liu, Hao and Pathak, Deepak},
  journal={arXiv preprint arXiv:2505.22660},
  year={2025}
}

@article{shenfeld2025rl,
  title={RL's Razor: Why Online Reinforcement Learning Forgets Less},
  author={Shenfeld, Idan and Pari, Jyothish and Agrawal, Pulkit},
  journal={arXiv preprint arXiv:2509.04259},
  year={2025}
}

@misc{zhang2025rlsurvey,
      title={A Survey of Reinforcement Learning for Large Reasoning Models}, 
      author={Kaiyan Zhang and Yuxin Zuo and Bingxiang He and Youbang Sun and Runze Liu and Che Jiang and Yuchen Fan and Kai Tian and Guoli Jia and Pengfei Li and Yu Fu and Xingtai Lv and Yuchen Zhang and Sihang Zeng and Shang Qu and Haozhan Li and Shijie Wang and Yuru Wang and Xinwei Long and Fangfu Liu and Xiang Xu and Jiaze Ma and Xuekai Zhu and Ermo Hua and Yihao Liu and Zonglin Li and Huayu Chen and Xiaoye Qu and Yafu Li and Weize Chen and Zhenzhao Yuan and Junqi Gao and Dong Li and Zhiyuan Ma and Ganqu Cui and Zhiyuan Liu and Biqing Qi and Ning Ding and Bowen Zhou},
      year={2025},
      eprint={2509.08827},
      archivePrefix={arXiv},
      primaryClass={cs.CL},
      url={https://arxiv.org/abs/2509.08827}, 
}

@article{rafailov2023direct,
  title={Direct preference optimization: Your language model is secretly a reward model},
  author={Rafailov, Rafael and Sharma, Archit and Mitchell, Eric and Manning, Christopher D and Ermon, Stefano and Finn, Chelsea},
  journal={Advances in neural information processing systems},
  volume={36},
  pages={53728--53741},
  year={2023}
}

@misc{qwen3,
      title={Qwen3 Technical Report}, 
      author={Qwen Team},
      year={2025},
      eprint={2505.09388},
      archivePrefix={arXiv},
      primaryClass={cs.CL},
      url={https://arxiv.org/abs/2505.09388}, 
}

@article{llama3,
  title={The llama 3 herd of models},
  author={Grattafiori, Aaron and Dubey, Abhimanyu and Jauhri, Abhinav and Pandey, Abhinav and Kadian, Abhishek and Al-Dahle, Ahmad and Letman, Aiesha and Mathur, Akhil and Schelten, Alan and Vaughan, Alex and others},
  journal={arXiv preprint arXiv:2407.21783},
  year={2024}
}

@article{octo,
  title={Octothinker: Mid-training incentivizes reinforcement learning scaling},
  author={Wang, Zengzhi and Zhou, Fan and Li, Xuefeng and Liu, Pengfei},
  journal={arXiv preprint arXiv:2506.20512},
  year={2025}
}

@misc{llmrl2025incorrect,
title={Incorrect Baseline Evaluations Call into Question Recent LLM-RL Claims},
author={Nikhil Chandak and Shashwat Goel and Ameya Prabhu},
year={2025},
howpublished={\url{https://safe-lip-9a8.notion.site/Incorrect-Baseline-Evaluations-Call-into-Question-Recent-LLM-RL-Claims-2012f1fbf0ee8094ab8ded1953c15a37?pvs=4}},
note={Notion Blog}
}

@misc{gpto4mini,
title={Introducing OpenAI o3 and o4-mini},
author={OpenAI},
year={2025},
url={https://openai.com/index/introducing-o3-and-o4-mini/}
}

@article{hendrycksmath2021,
  title={Measuring Mathematical Problem Solving With the MATH Dataset},
  author={Dan Hendrycks and Collin Burns and Saurav Kadavath and Akul Arora and Steven Basart and Eric Tang and Dawn Song and Jacob Steinhardt},
  journal={NeurIPS},
  year={2021}
}

@inproceedings{lightman2023let,
  title={Let's verify step by step},
  author={Lightman, Hunter and Kosaraju, Vineet and Burda, Yuri and Edwards, Harrison and Baker, Bowen and Lee, Teddy and Leike, Jan and Schulman, John and Sutskever, Ilya and Cobbe, Karl},
  booktitle={The Twelfth International Conference on Learning Representations},
  year={2023}
}

@inproceedings{rein2024gpqa,
  title={Gpqa: A graduate-level google-proof q\&a benchmark},
  author={Rein, David and Hou, Betty Li and Stickland, Asa Cooper and Petty, Jackson and Pang, Richard Yuanzhe and Dirani, Julien and Michael, Julian and Bowman, Samuel R},
  booktitle={First Conference on Language Modeling},
  year={2024}
}

@article{hendrycks2020measuring,
  title={Measuring massive multitask language understanding},
  author={Hendrycks, Dan and Burns, Collin and Basart, Steven and Zou, Andy and Mazeika, Mantas and Song, Dawn and Steinhardt, Jacob},
  journal={arXiv preprint arXiv:2009.03300},
  year={2020}
}

@article{wang2024mmlu,
  title={Mmlu-pro: A more robust and challenging multi-task language understanding benchmark},
  author={Wang, Yubo and Ma, Xueguang and Zhang, Ge and Ni, Yuansheng and Chandra, Abhranil and Guo, Shiguang and Ren, Weiming and Arulraj, Aaran and He, Xuan and Jiang, Ziyan and others},
  journal={Advances in Neural Information Processing Systems},
  volume={37},
  pages={95266--95290},
  year={2024}
}

@inproceedings{kwon2023efficient,
  title={Efficient Memory Management for Large Language Model Serving with PagedAttention},
  author={Woosuk Kwon and Zhuohan Li and Siyuan Zhuang and Ying Sheng and Lianmin Zheng and Cody Hao Yu and Joseph E. Gonzalez and Hao Zhang and Ion Stoica},
  booktitle={Proceedings of the ACM SIGOPS 29th Symposium on Operating Systems Principles},
  year={2023}
}

@article{sheng2024hybridflow,
  title   = {HybridFlow: A Flexible and Efficient RLHF Framework},
  author  = {Guangming Sheng and Chi Zhang and Zilingfeng Ye and Xibin Wu and Wang Zhang and Ru Zhang and Yanghua Peng and Haibin Lin and Chuan Wu},
  year    = {2024},
  journal = {arXiv preprint arXiv: 2409.19256}
}

@misc{yuan2023scaling,
      title={Scaling Relationship on Learning Mathematical Reasoning with Large Language Models}, 
      author={Zheng Yuan and Hongyi Yuan and Chengpeng Li and Guanting Dong and Keming Lu and Chuanqi Tan and Chang Zhou and Jingren Zhou},
      year={2023},
      eprint={2308.01825},
      archivePrefix={arXiv},
      primaryClass={cs.CL},
      url={https://arxiv.org/abs/2308.01825}, 
}

@misc{openr1,
    title = {Open R1: A fully open reproduction of DeepSeek-R1},
    url = {https://github.com/huggingface/open-r1},
    author = {Hugging Face},
    month = {January},
    year = {2025}
}

@misc{bercovich2025llamanemotron,
      title={Llama-Nemotron: Efficient Reasoning Models}, 
      author={Akhiad Bercovich and Itay Levy and Izik Golan and Mohammad Dabbah and Ran El-Yaniv and Omri Puny and Ido Galil and Zach Moshe and Tomer Ronen and Najeeb Nabwani and Ido Shahaf and Oren Tropp and Ehud Karpas and Ran Zilberstein and Jiaqi Zeng and Soumye Singhal and Alexander Bukharin and Yian Zhang and Tugrul Konuk and Gerald Shen and Ameya Sunil Mahabaleshwarkar and Bilal Kartal and Yoshi Suhara and Olivier Delalleau and Zijia Chen and Zhilin Wang and David Mosallanezhad and Adi Renduchintala and Haifeng Qian and Dima Rekesh and Fei Jia and Somshubra Majumdar and Vahid Noroozi and Wasi Uddin Ahmad and Sean Narenthiran and Aleksander Ficek and Mehrzad Samadi and Jocelyn Huang and Siddhartha Jain and Igor Gitman and Ivan Moshkov and Wei Du and Shubham Toshniwal and George Armstrong and Branislav Kisacanin and Matvei Novikov and Daria Gitman and Evelina Bakhturina and Jane Polak Scowcroft and John Kamalu and Dan Su and Kezhi Kong and Markus Kliegl and Rabeeh Karimi and Ying Lin and Sanjeev Satheesh and Jupinder Parmar and Pritam Gundecha and Brandon Norick and Joseph Jennings and Shrimai Prabhumoye and Syeda Nahida Akter and Mostofa Patwary and Abhinav Khattar and Deepak Narayanan and Roger Waleffe and Jimmy Zhang and Bor-Yiing Su and Guyue Huang and Terry Kong and Parth Chadha and Sahil Jain and Christine Harvey and Elad Segal and Jining Huang and Sergey Kashirsky and Robert McQueen and Izzy Putterman and George Lam and Arun Venkatesan and Sherry Wu and Vinh Nguyen and Manoj Kilaru and Andrew Wang and Anna Warno and Abhilash Somasamudramath and Sandip Bhaskar and Maka Dong and Nave Assaf and Shahar Mor and Omer Ullman Argov and Scot Junkin and Oleksandr Romanenko and Pedro Larroy and Monika Katariya and Marco Rovinelli and Viji Balas and Nicholas Edelman and Anahita Bhiwandiwalla and Muthu Subramaniam and Smita Ithape and Karthik Ramamoorthy and Yuting Wu and Suguna Varshini Velury and Omri Almog and Joyjit Daw and Denys Fridman and Erick Galinkin and Michael Evans and Katherine Luna and Leon Derczynski and Nikki Pope and Eileen Long and Seth Schneider and Guillermo Siman and Tomasz Grzegorzek and Pablo Ribalta and Monika Katariya and Joey Conway and Trisha Saar and Ann Guan and Krzysztof Pawelec and Shyamala Prayaga and Oleksii Kuchaiev and Boris Ginsburg and Oluwatobi Olabiyi and Kari Briski and Jonathan Cohen and Bryan Catanzaro and Jonah Alben and Yonatan Geifman and Eric Chung and Chris Alexiuk},
      year={2025},
      eprint={2505.00949},
      archivePrefix={arXiv},
      primaryClass={cs.CL},
      url={https://arxiv.org/abs/2505.00949}, 
}

@misc{AIME,
    title = {AIME. AIME problems and solutions, 2024.},
    author = {AIME},
    year = {2024},
    url = {https://artofproblemsolving.com/wiki/index.php/American_Invitational_Mathematics_Examination}
}

@article{srivastava2023beyond,
  title={Beyond the Imitation Game: Quantifying and extrapolating the capabilities of language models},
  author={BIG-bench authors},
  journal={Transactions on Machine Learning Research},
  issn={2835-8856},
  year={2023},
  url={https://openreview.net/forum?id=uyTL5Bvosj},
  note={}
}

@article{suzgun2022challenging,
  title={Challenging BIG-Bench Tasks and Whether Chain-of-Thought Can Solve Them},
  author={Suzgun, Mirac and Scales, Nathan and Sch{\"a}rli, Nathanael and Gehrmann, Sebastian and Tay, Yi and Chung, Hyung Won and Chowdhery, Aakanksha and Le, Quoc V and Chi, Ed H and Zhou, Denny and and Wei, Jason},
  journal={arXiv preprint arXiv:2210.09261},
  year={2022}
}

@article{lu2025arpo,
  title={ARPO: End-to-End Policy Optimization for GUI Agents with Experience Replay},
  author={Lu, Fanbin and Zhong, Zhisheng and Liu, Shu and Fu, Chi-Wing and Jia, Jiaya},
  journal={arXiv preprint arXiv:2505.16282},
  year={2025}
}

@article{cheng2025reasoning,
  title={Reasoning with exploration: An entropy perspective},
  author={Cheng, Daixuan and Huang, Shaohan and Zhu, Xuekai and Dai, Bo and Zhao, Wayne Xin and Zhang, Zhenliang and Wei, Furu},
  journal={arXiv preprint arXiv:2506.14758},
  year={2025}
}

@article{bartoldson2025trajectory,
  title={Trajectory balance with asynchrony: Decoupling exploration and learning for fast, scalable llm post-training},
  author={Bartoldson, Brian R and Venkatraman, Siddarth and Diffenderfer, James and Jain, Moksh and Ben-Nun, Tal and Lee, Seanie and Kim, Minsu and Obando-Ceron, Johan and Bengio, Yoshua and Kailkhura, Bhavya},
  journal={arXiv preprint arXiv:2503.18929},
  year={2025}
}

@article{balunovic2025matharena,
  title = {MathArena: Evaluating LLMs on Uncontaminated Math Competitions},
  author = {Mislav Balunović and Jasper Dekoninck and Ivo Petrov and Nikola Jovanović and Martin Vechev},
  journal = {Proceedings of the Neural Information Processing Systems Track on Datasets and Benchmark},
  year={2025}
}

@misc{sun2025challengingboundariesreasoningolympiadlevel,
      title={Challenging the Boundaries of Reasoning: An Olympiad-Level Math Benchmark for Large Language Models},
      author={Haoxiang Sun and Yingqian Min and Zhipeng Chen and Wayne Xin Zhao and Zheng Liu and Zhongyuan Wang and Lei Fang and Ji-Rong Wen},
      year={2025},
      eprint={2503.21380},
      archivePrefix={arXiv},
      primaryClass={cs.CL},
      url={https://arxiv.org/abs/2503.21380},
}

@article{liu2025simple,
  title={A Simple" Try Again" Can Elicit Multi-Turn LLM Reasoning},
  author={Liu, Licheng and Wang, Zihan and Li, Linjie and Xu, Chenwei and Lu, Yiping and Liu, Han and Sil, Avirup and Li, Manling},
  journal={arXiv preprint arXiv:2507.14295},
  year={2025}
}

@article{zhang2025critique,
  title={Critique-grpo: Advancing llm reasoning with natural language and numerical feedback},
  author={Zhang, Xiaoying and Sun, Hao and Zhang, Yipeng and Feng, Kaituo and Lu, Chaochao and Yang, Chao and Meng, Helen},
  journal={arXiv preprint arXiv:2506.03106},
  year={2025}
}

@article{li2025questa,
  title={Questa: Expanding reasoning capacity in llms via question augmentation},
  author={Li, Jiazheng and Lin, Hongzhou and Lu, Hong and Wen, Kaiyue and Yang, Zaiwen and Gao, Jiaxuan and Wu, Yi and Zhang, Jingzhao},
  journal={arXiv preprint arXiv:2507.13266},
  year={2025}
}

@article{qu2025rlad,
  title={RLAD: Training LLMs to Discover Abstractions for Solving Reasoning Problems},
  author={Qu, Yuxiao and Singh, Anikait and Lee, Yoonho and Setlur, Amrith and Salakhutdinov, Ruslan and Finn, Chelsea and Kumar, Aviral},
  journal={arXiv preprint arXiv:2510.02263},
  year={2025}
}

@article{wang2025dissmath,
  title={Beyond Accuracy: Dissecting Mathematical Reasoning for LLMs Under Reinforcement Learning},
  author={Wang, Jiayu and Ming, Yifei and Ke, Zixuan and Xiong, Caiming and Joty, Shafiq and Albarghouthi, Aws and Sala, Frederic},
  journal={arXiv preprint arXiv:2506.04723},
  year={2025}
}
